\documentclass[11pt]{article}

\usepackage[final]{acl}

\usepackage{times}
\usepackage{latexsym}

\usepackage[T1]{fontenc}

\usepackage[utf8]{inputenc}

\usepackage{microtype}

\usepackage{inconsolata}

\usepackage{graphicx}
\usepackage{amsmath,amssymb}
\usepackage{booktabs,array,adjustbox,colortbl}
\definecolor{TableGroup}{gray}{0.93}
\definecolor{OursTint}{RGB}{205,226,237}
\newcommand{\HThree}{Minimax-H3}
\newcommand{\OurMethod}{ \textbf{PhyEvolve}}
\usepackage{times}
\usepackage{latexsym}

\usepackage{xcolor}
\usepackage{tcolorbox}
\tcbuselibrary{skins}

\definecolor{boxTitleBg}{RGB}{80, 120, 160}    
\definecolor{boxContentBg}{RGB}{240, 245, 250} 
\definecolor{boxFrameColor}{RGB}{80, 120, 160} 

\newtcolorbox{systemprompt}[1][]{
  enhanced,               
  title={System Instruction}, 
  colframe=boxFrameColor, 
  colbacktitle=boxTitleBg,
  coltitle=white,         
  colback=boxContentBg,   
  fonttitle=\bfseries\rmfamily, 
  fontupper=\rmfamily,    
  arc=1.5mm,              
  boxrule=0.7pt,          
  titlerule=0pt,          
  left=5pt, right=5pt, top=5pt, bottom=5pt, 
  #1 
}

\usepackage[T1]{fontenc}

\usepackage[utf8]{inputenc}

\usepackage{microtype}

\usepackage{inconsolata}

\usepackage{graphicx}

\title{PhyEvolve: Multi-Agent State-Evolution Planning for Physics-Aware Video Generation}

\author{
  Yang-Hao Zhou$^{\dagger}$ \quad
  Lixing Tan$^{\dagger}$ \quad
  Zhiqi Li \quad
  Guang Yang \quad
  Xiao Yang$^{*,\ddagger}$ \\
  Canva Research \\
  $^{\dagger}$Equal contribution \\
  $^{*}$Corresponding author \\
  $^{\ddagger}$Project leader
}

\begin{document}
\maketitle

\begin{abstract}

Recent advances in text-to-video generation have substantially improved visual fidelity, yet existing models still struggle to understand the physical evolution implicit in user prompts, making it difficult to generate videos with continuous physical state evolution. Here, we propose PhyEvolve, a multi-agent framework that enhances natural-language prompts through structured state-evolution planning. PhyEvolve consists of three agents: a Director, a Physical Evolver, and a Composer. Given a user prompt and an optional first frame, the Director grounds the scene and disentangles physical events from observation changes. The Physical Evolver uses scene evidence to identify physical mechanisms then constructs causally linked state transitions that unfold throughout the video, whether visible or not. The Composer utilizes these transitions to enhance the original prompts. Without retraining or model modification, PhyEvolve works directly with existing generators through prompt replacement. Extensive experiments on four major benchmarks show that PhyEvolve achieves state-of-the-art performance, outperforming the cutting-edge prior agentic baseline by 7.9 percentage points on average. These results demonstrate that explicit causal state-evolution planning improves physical consistency and state continuity in generated videos through natural-language prompt enhancement.

\end{abstract}

\section{Introduction}

\begin{figure*}[!t]
  \centering
  \includegraphics[width=\textwidth]{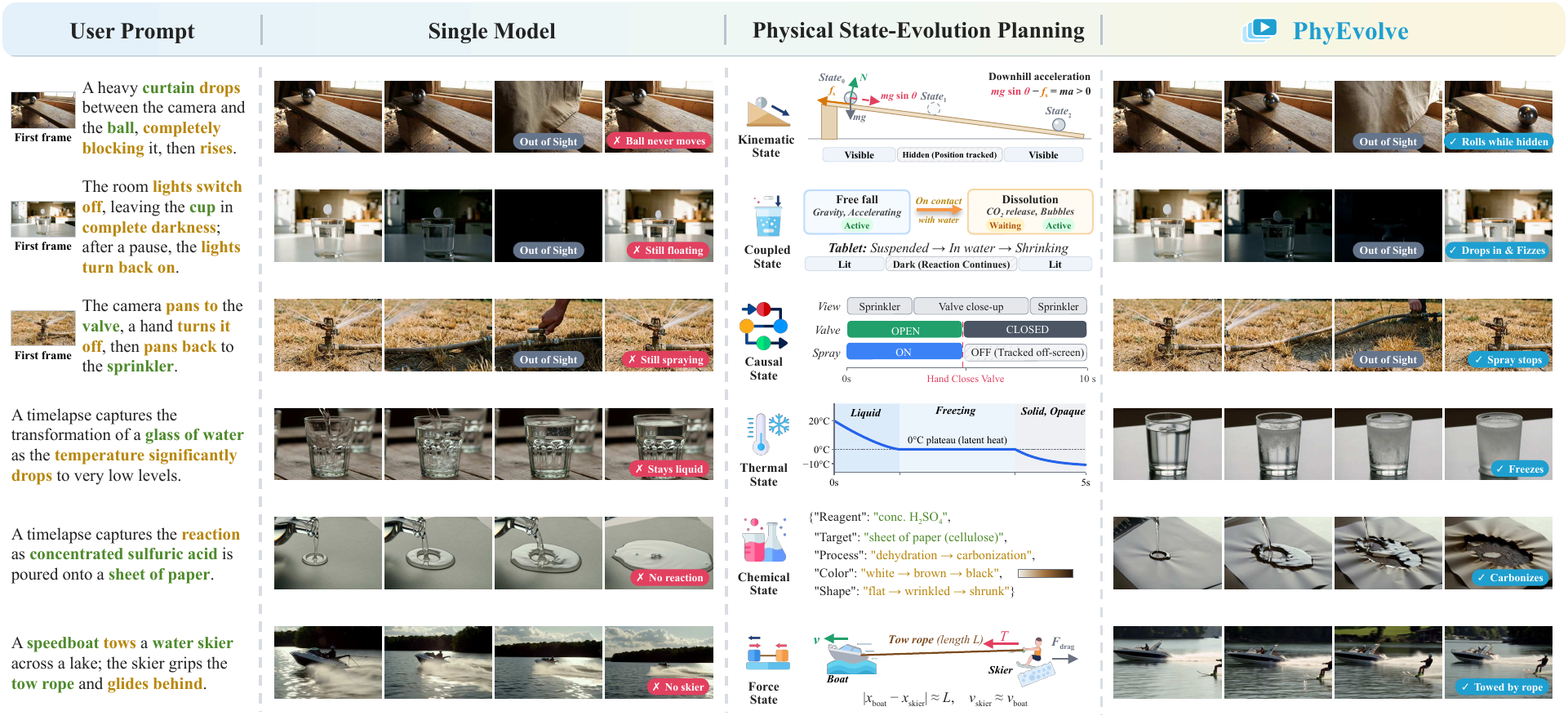}
  \caption{\textbf{PhyEvolve plans physical state evolution before video generation.}
  Compared with single-pass Minimax-H3, PhyEvolve produces physically consistent videos,
  including when processes continue out of sight.}
  \label{fig:teaser}
\end{figure*}

Physics-aware video generation requires scenes to evolve consistently with the conditions and interactions described by users. Existing models remain unreliable in this respect~\citep{meng2024towards,bansal2026videophy}. Figure~\ref{fig:teaser} illustrates failures to freeze water, reproduce material reactions, stop spraying after valve closure, or depict the skier towed by a boat. Brief prompts often state the scene and actions while leaving the mechanisms, activation conditions, and resulting state changes implicit. These failures motivate prompt enhancement that makes physical consequences explicit while preserving the requested objects, actions, and interactions.

Observation changes make this task particularly challenging~\citep{ma2026out}. In the sprinkler example, camera movement changes visibility, whereas closing the valve changes the conditions governing water flow. The sprinkler's state must reflect this intervention even while it is out of view. Enhancement must therefore distinguish an event's physical effects from its effects on visibility and propagate states under conditions that remain active or are altered by events. The inferred outcome must also reach the final prompt: merely describing the camera's return leaves the sprinkler's expected state unstated.

Prior work enriches generation conditions through physical reasoning, tool orchestration, and causal event planning~\citep{xue2025phyt2v,feng2026newton,wang2026physically}. We focus on preserving event effects and state dependencies throughout interpretation, planning, and prompt composition. We formulate this task as semantic completion through structured state-evolution planning. Explicit records connect scene evidence, mechanism activation conditions, and dependent state transitions, linking the original instruction to the final prompt. The same records serve two purposes: they support reasoning about how the scene evolves and provide a basis for checking whether the final text preserves the planned physical changes.

We introduce PhyEvolve, a multi-agent framework that implements this formulation. Given a user instruction and an optional initial frame, the Director identifies scene entities and their initial states from textual and visual evidence. It extracts events and specifies which physical conditions, object states, or aspects of visibility each event is permitted to affect. The Physical Evolver then organizes local physical records around evidence, conditions, mechanisms, and responses, and uses them to infer dependent state transitions. Each inferred state becomes the starting point for subsequent reasoning, so evolution across observed and unobserved intervals builds on preceding physical effects. The resulting evolution records specify the object states expected in subsequent events while retaining uncertainty where the input does not determine a change.

The Composer carries these planned changes into the text submitted to the generator. It integrates the original requirements, the Director’s plan and constraints, and the Physical Evolver’s records into a complete replacement prompt. It checks the draft against these records for omissions and conflicts, incorporates physical conditions and state changes that have not been expressed, and revises only the affected sentences before checking them again. This process allows the physical reasoning to be retained in the text submitted to the generator while constraining revisions to mechanisms supported by the upstream records. Each agent's output is reviewed and local revision before final generation. The enhanced prompt then replaces the original prompt and is passed to the video generator together with the optional initial frame, without retraining or modifying the generator.

We evaluate the downstream benefits of enhanced prompts on StEvo-Bench~\citep{ma2026out}, PhyGenBench~\citep{meng2024towards}, PhyWorldBench~\citep{gu2026phyworldbench}, and VideoPhy-2~\citep{bansal2026videophy}, covering state evolution under interrupted observation and physical commonsense across diverse physical laws, object interactions, and complex actions. With Minimax-H3~\cite{minimaxh32026} retained as the generation backend, PhyEvolve outperforms direct generation and the H3-based versions of NEWTON~\citep{feng2026newton} and Co-Director~\citep{song2026codirector} on the primary metric of each benchmark. Averaged across the four primary benchmark metrics, PhyEvolve gains 14.9 percentage points over direct generation and 7.9 points over NEWTON, the prior agentic baseline with the highest mean score. On StEvo-Bench, Evolution Success increases from 17.7\% under direct generation to 45.7\% with PhyEvolve. These results support explicit state planning and faithful prompt composition as a means of improving physical consistency and state continuity in existing video generators.

Our main contributions are as follows:

\begin{itemize}

\item We formulate prompt enhancement for physics-aware video generation as semantic completion through structured state-evolution planning, distinguishing the physical effects of events from their effects on observation and accounting for state progression during unobserved intervals.

\newcommand{\method}{PhyEvolve}

\begin{figure*}[!t]
  \centering
  \includegraphics[width=\textwidth]{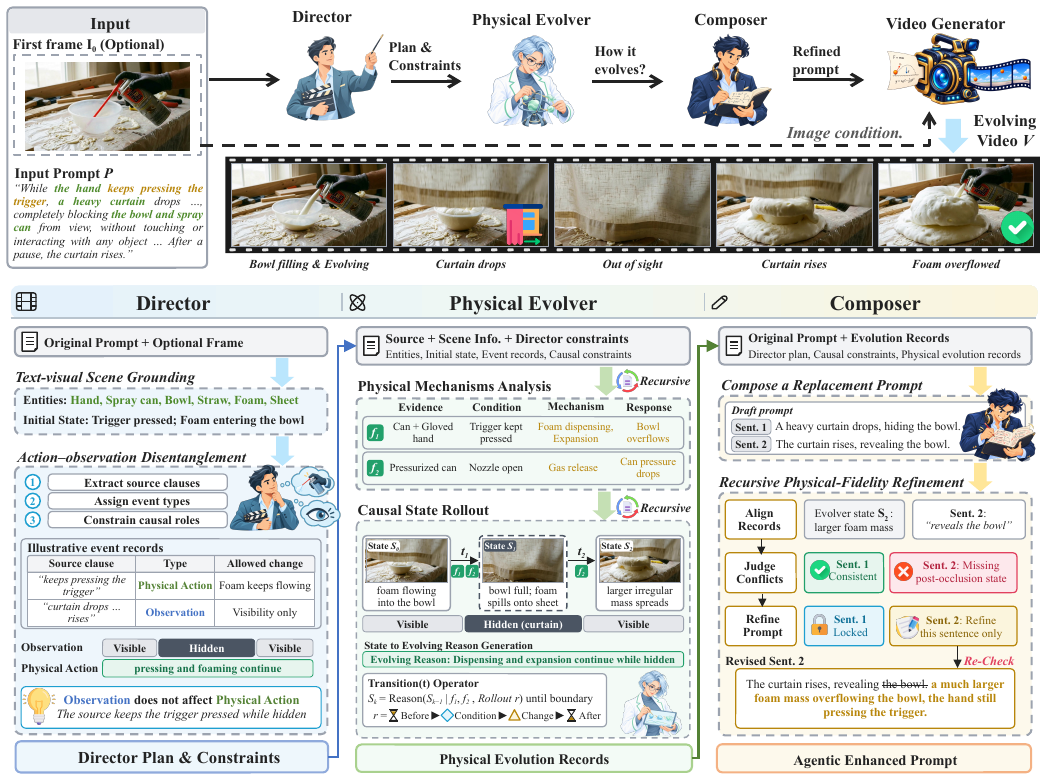}

\caption{Overview of \method{}. The Director separates actions from observations, the Physical Evolver models causal state evolution under occlusion, and the Composer refines prompts for physically consistent video generation.}
  \label{fig:overview}
\end{figure*}

\item We propose PhyEvolve, in which a Director, a Physical Evolver, and a Composer connect event interpretation, physical mechanisms analysis, causal state rollout, and prompt composition. Grounded intermediate records, review, and local revision at each stage preserve the original intent and inferred evolution.

\item We evaluate enhanced prompts on four physical video benchmarks using a frozen H3 backend, demonstrating improvements in physical consistency without retraining or modifying the video generator.
\end{itemize}

\section{Related Work}
\label{sec:related_work}

\subsection{Physics-Aware Video Generation}
\label{sec:related_physics_video}

Research on physical scene dynamics ranges from object-centric intuitive physics~\citep{piloto2022intuitive} to latent dynamics for prediction and planning~\citep{hafner2019learning}. Video world models address prediction and long-horizon generation~\citep{agarwal2025cosmos,yuan2026helios,huang2026solarwm}, with memory and geometric consistency supporting longer rollouts~\citep{wu2026infinite,sun2025worldplay}. Memory-based methods retain spatiotemporal context~\citep{stapf2026composition}, track hidden dynamic subjects~\citep{chen2026out}, and learn state evolution across observation interruptions~\citep{xu2026teaching}. Evaluation spans general video quality~\citep{huang2024vbench} and compositionality~\citep{sun2025t2v}, physical commonsense~\citep{meng2024towards,gu2026phyworldbench,bansal2025videophy,bansal2026videophy}, agreement with observed or controlled dynamics~\citep{motamed2026generative,tragoudaras2025evaluating,li2026learning}, and out-of-sight state evolution~\citep{ma2026out}. PhyEvolve focuses on explicitly planning how physical states change across visible and unobserved intervals.

\subsection{Agentic Video Generation}
\label{sec:related_agentic_video}

Agentic video generation combines language-model planning, tool coordination, and refinement. VideoDirectorGPT organizes scenes and entity layouts~\citep{lin2023videodirectorgpt}, while multi-agent systems coordinate generative tools~\citep{yuan2024mora}, storytelling stages~\citep{hu2024storyagent,wang2026mavis}, and compositional or narrative refinement~\citep{huang2024genmac,song2026codirector}. Executable event graphs~\citep{cudlenco2026agentic} and Code World Model~\citep{chen2026code} further connect plans to executable representations. Physics-oriented frameworks offer closer comparisons: NEWTON trains a planner to coordinate physics-aware tools and iterative verification around an unchanged video generator~\citep{feng2026newton}; MoReGen combines multi-agent reasoning with executable physics simulation and rendering~\citep{bai2026moregen}. PhyEvolve centers collaboration on a causal state-evolution plan. Its agents distinguish physical actions from observation changes, infer local mechanisms and dependent state transitions, and express their consequences in a complete natural-language prompt. Stage-wise review and local repair occur before generation, supporting physical evolution through unobserved intervals without retraining or modifying the generator.

\section{Method}
\label{sec:method}

\subsection{Overview}
\label{sec:overview}

Given a user prompt $P$ and an optional initial frame $I_0$, PhyEvolve constructs an enhanced replacement prompt through three cooperating agents. As shown in Figure~\ref{fig:overview}, the Director grounds the scene and assigns causal roles to input events; the Physical Evolver identifies applicable mechanisms and rolls them forward into a state trajectory; and the Composer expresses the reviewed records as a complete video-generation prompt. The workflow is summarized as
\begin{equation}
\begin{aligned}
\mathcal{D} &= \operatorname{Director}(P,I_0),\\
(\mathcal{M},\mathcal{T}) &= \operatorname{Evolver}(P,\mathcal{D}),\\
P^{\star} &= \operatorname{Composer}(P,\mathcal{D},\mathcal{M},\mathcal{T}),\\
V &= G(P^{\star},I_0),
\end{aligned}
\label{eq:pipeline}
\end{equation}

where $\mathcal{D}$ is the Director plan, $\mathcal{M}$ contains local mechanism records, $\mathcal{T}$ is the state-evolution trajectory, and $G$ is the unchanged video generator. Each agent reviews its own structured output and locally revises failed entries. All review and repair occur before $G$ is invoked; the generated video is not fed back into the agent loop.

\subsection{Director Agent}
\label{sec:director_planning}

\paragraph{Text--visual scene grounding.}
The Director first extracts entities, initial states, spatial relations, and their textual or visual evidence from $P$ and $I_0$. This grounded scene establishes the starting conditions without reconstructing their prior history or predicting later physical outcomes.

\paragraph{Action--observation disentanglement.}
It then locates the source clauses describing relevant events and classifies them as physical actions, environmental changes, observation changes, or scene constraints. For each event, the Director records whether it may alter physical conditions, object state, visibility, or a combination of them. An observation-only event, such as a non-contact curtain used solely to hide an object, changes visibility but does not itself initiate, interrupt, or reset physical evolution. If the input explicitly assigns a physical effect to an event, that effect is retained. A local review checks grounding fidelity, event types, and causal roles before passing the resulting scene plan and constraints $\mathcal{D}$ to the Physical Evolver.

\subsection{Physical Evolver Agent}
\label{sec:physical_evolution}

The Physical Evolver receives $P$ and the reviewed Director output. It performs physical mechanisms analysis followed by causal state rollout. A separate call jointly reviews these records.

\paragraph{Physical mechanisms analysis.}
The Evolver represents each supported mechanism as
\begin{equation}
\mathcal{M}=\{m_i\}_{i=1}^{N},
\qquad
m_i=(e_i,c_i,\phi_i,r_i),
\label{eq:local_mechanisms}
\end{equation}

where $e_i$, $c_i$, $\phi_i$, and $r_i$ denote the supporting evidence, activation condition, physical mechanism, and expected response. These mechanism records provide the basis reasoning for subsequent causal state rollout within the same generation call.

\paragraph{Causal state rollout.}

The identified mechanisms are organized into dependent transitions. Each transition records the before-state, active conditions, change process, after-state, supporting mechanism IDs, and dependencies on earlier transitions. Physical state and visibility are represented separately:
\begin{equation}
s_k \xrightarrow[c_k]{\mathcal{M}_k} s_{k+1},
\qquad
o_k=\mathcal{O}(s_k,v_k),
\label{eq:causal_rollout}
\end{equation}

where $s_k$ is the physical state, $c_k$ contains the active conditions, $\mathcal{M}_k\subseteq\mathcal{M}$ contains the applicable mechanisms, and $o_k$ is what can be observed under visibility condition $v_k$. The index $k$ denotes causal order rather than a calibrated timestamp. Each after-state is carried into subsequent transitions. Therefore, when an event changes only $v_k$, mechanisms whose conditions remain active continue during occlusion, and reappearance is described from the evolved state. 
A single separate review then jointly checks mechanism grounding, conditions and responses, state continuity, transition dependencies, and consistency with the Director constraints.

\begin{table*}[!t]
\centering
\caption{Unified evaluation of video generators and
agentic frameworks across physical-video benchmarks.
All scores are reported as percentages; higher is better.
Average denotes the unweighted arithmetic mean across the four benchmarks.}
\label{tab:unified_main}
\small
\setlength{\tabcolsep}{4pt}
\renewcommand{\arraystretch}{1.12}
\begin{adjustbox}{max width=\textwidth}
\begin{tabular}{llccccc}
\toprule
Method & Type 
& \shortstack{StEvo-Bench\\Evo. Success $\uparrow$}
& \shortstack{PhyGenBench\\Average $\uparrow$}
& \shortstack{PhyWorldBench\\Overall Both $\uparrow$}
& \shortstack{VideoPhy-2\\All Joint $\uparrow$}
& Average $\uparrow$ \\
\midrule

\multicolumn{7}{l}{\textit{Open-source video generators}} \\
Wan2.2-TI2V-5B
    & Video Generator & 0.9 & 66.7 & 15.0 & 24.9 & 26.9 \\
LTX-Video-2.5
    & Video Generator & 0.9 & 64.6 & 16.5 & 22.2 & 26.1 \\
Minimax-H3 
    & Video Generator & 17.7 & 63.5 & 66.6 & 23.9 & 42.9 \\

\midrule
\multicolumn{7}{l}{\textit{H3-based systems}} \\
NEWTON
    & Agentic Framework & 29.0 & 77.5 & 69.9 & 23.2 & 49.9 \\
Co-Director
    & Agentic Framework & 20.0 & 75.4 & 65.8 & 24.9 & 46.5 \\
\rowcolor{OursTint}
\textbf{PhyEvolve}
    & Agentic Framework
    & \textbf{45.7} & \textbf{79.6}
    & \textbf{75.8} & \textbf{30.1} & \textbf{57.8} \\

\bottomrule
\end{tabular}
\end{adjustbox}
\end{table*}

\begin{table}[!t]
\centering

\caption{StEvo-Bench: state evolution and control metrics.}
\label{tab:06_stevo}
\label{tab:13_stevo_controls}
\begingroup
\footnotesize
\setlength{\tabcolsep}{3pt}
\renewcommand{\arraystretch}{1.12}
\begin{adjustbox}{max width=\linewidth}
\begin{tabular}{@{}l*{6}{c}@{}}
\toprule
Model
& \shortstack{Evo.\\Success $\uparrow$}
& Prog. $\uparrow$
& Phys. $\uparrow$
& Coh. $\uparrow$
& \shortstack{Obs.\\Ctrl. $\uparrow$}
& \shortstack{Act.\\Ctrl. $\uparrow$} \\
\midrule

\multicolumn{7}{l}{\textit{Open-source video generators}} \\
Wan2.2-TI2V-5B
    & 0.9 & 10.9 & 51.6 & 62.4 & 45.7 & 54.3 \\
LTX-Video-2.5
    & 0.9 & 14.0 & 66.5 & \textbf{95.0} & 55.9 & 56.8 \\
\HThree
    & 17.7 & 25.3 & 70.6 & 88.2 & 87.6 & 81.5 \\

\midrule

\multicolumn{7}{l}{\textit{H3-based agentic frameworks}} \\
NEWTON
    & 29.0 & 36.8 & 93.4 & 93.2 & 89.3 & 80.7 \\
Co-Director
    & 20.0 & 42.0 & 72.0 & 94.0 & 83.7 & 83.3 \\

\rowcolor{OursTint}
\OurMethod
    & \textbf{45.7} & \textbf{51.2} & \textbf{96.8} & 94.6 & \textbf{89.5} & \textbf{85.2} \\

\bottomrule
\end{tabular}
\end{adjustbox}
\par\endgroup
\end{table}

\begin{table}[!t]
\centering

\caption{PhyGenBench: physical commonsense accuracy.}
\label{tab:02_phygenbench}

\begingroup
\small
\setlength{\tabcolsep}{4pt}
\renewcommand{\arraystretch}{1.15}
\begin{adjustbox}{max width=\columnwidth}
\begin{tabular}{l*{5}{c}}
\toprule

Model
& Mechanics $\uparrow$ & Optics $\uparrow$ & Thermal $\uparrow$
& Material $\uparrow$ & Average $\uparrow$ \\
\midrule

\multicolumn{6}{l}{\textit{Open-source video generators}} \\
\addlinespace[2pt]
CogVideoX-2B       & 38.0 & 43.0 & 34.0 & 39.0 & 39.0 \\
CogVideoX-5B       & 39.0 & 55.0 & 40.0 & 42.0 & 45.0 \\
Open-Sora V1.2  & 43.0 & 50.0 & 44.0 & 37.0 & 44.0 \\
Lavie          & 30.0 & 44.0 & 38.0 & 32.0 & 36.0 \\
Vchitect 2.0   & 41.0 & 56.0 & 44.0 & 37.0 & 45.0 \\
Wan2.2-TI2V-5B & 66.7 & 82   & 57.8 & 54.2 & 66.7 \\
LTX-Video-2.5  & 64.2 & 70.7 & 58.9 & 61.7 & 64.6 \\
\HThree        & 59.2 & 78.0 & 57.8 & 54.2 & 63.5 \\

\midrule
\multicolumn{6}{l}{\textit{Closed-source video generators}} \\
\addlinespace[2pt]
Pika           & 35.0 & 56.0 & 43.0 & 39.0 & 44.0 \\
Gen-3          & 45.0 & 57.0 & 49.0 & 51.0 & 51.0 \\
Kling          & 45.0 & 58.0 & 50.0 & 40.0 & 49.0 \\

\midrule

\multicolumn{6}{l}{\textit{H3-based Agentic frameworks}} \\
\addlinespace[2pt]
NEWTON         & 70.0 & \textbf{89.3} & 76.7 & 70.8 & 77.5 \\
Co-Director    & 69.2 & 80.7  & 76.7  & 74.2  & 75.4 \\
\rowcolor{OursTint}
\OurMethod     & \textbf{73.3} & 84.7  & \textbf{82.2}  & \textbf{77.5}  & \textbf{79.6} \\
\bottomrule
\end{tabular}
\end{adjustbox}
\endgroup

\end{table}

\subsection{Composer Agent}
\label{sec:prompt_composition}

\paragraph{Replacement prompt composition.}
The Composer combines $P$, $\mathcal{D}$, $\mathcal{M}$ and $\mathcal{T}$ into a self-contained replacement prompt. The draft preserves the requested entities, actions, event order, scene constraints, and pacing while expressing the physical conditions and state changes needed to connect successive events.

\paragraph{Recursive physical-fidelity refinement.}
Physical-fidelity refinement aligns the draft with the upstream records and flags sentences that omit a required state, contradict a causal constraint, or alter the original request. The Composer revises only the flagged sentences and keeps accepted sentences fixed. It may verbalize mechanisms already present in $\mathcal{T}$ but cannot introduce new physical mechanisms. Refinement stops when the prompt passes review or reaches the retry budget; if final validation still fails, the system falls back to the last enhanced prompt. The selected prompt $P^{\star}$, together with the same initial frame when the task is image-conditioned, is then supplied to $G$. For text-only tasks, the agents plan using textual evidence alone and supply the enhanced prompt to the video generator.

\section{Experiments}

\subsection{Benchmark Datasets}
\label{subsec:benchmark_dataset}

We assess state evolution under interrupted observation and broader physical correctness using four benchmarks. PhyGenBench~\citep{meng2024towards} contains 160 prompts covering 27 physical laws in mechanics, optics, thermal phenomena, and material properties; we report its benchmark score (\emph{Average}). PhyWorldBench~\citep{gu2026phyworldbench} contains 1,050 prompts across 10 categories and 50 subcategories, with scenarios covering fundamental, composite, and deliberately unphysical phenomena. We report \emph{Overall Both}, which requires semantic adherence and satisfaction of the scenario-specific physical evaluation criteria. VideoPhy-2~\citep{bansal2026videophy} contains 3,940 prompts derived from 197 real-world actions, including 590 prompts in its original test split. We report automatic \emph{All Joint} and \emph{Hard Joint} scores under the NEWTON reporting protocol; Joint requires both semantic adherence and physical commonsense scores to reach at least 4 on five-point scales. StEvo-Bench~\citep{ma2026out} contains 225 tasks across six categories, pairing initial images with action prompts and observation interruptions through occlusion, illumination changes, or camera movement. Its \emph{Evolution Success} metric requires state progress, physical plausibility, and temporal coherence on videos that pass both action and observation control.

\begin{table}[!t]
\centering
\caption{PhyWorldBench: physics-category results.
All scores are reported for the \textit{Both} metric.
}
\label{tab:03_phyworldbench}

\begingroup
\footnotesize
\setlength{\tabcolsep}{3.2pt}
\renewcommand{\arraystretch}{1.15}
\begin{adjustbox}{max width=\columnwidth}
\begin{tabular}{lcccc}
\toprule
Model
& \shortstack{Fundamental\\Physics}
& \shortstack{Composite\\Physics}
& Anti-Physics
& \shortstack{Overall\\Performance} \\
\midrule

\multicolumn{5}{l}{\textit{Closed-source video generators}} \\
Sora-Turbo        & 24.6 & 18.8 & 3.9 & 20.8 \\
Gen-3             & 16.1 & 10.3 & 2.0 & 13.0 \\
Kling-1.6         & 23.5 & 13.9 & 4.2 & 18.8 \\
Pika 2.0          & 31.2 & 23.9 & 1.1 & 26.2 \\
Luma              & 22.0 & 16.0 & 0.0 & 18.4 \\

\midrule

\multicolumn{5}{l}{\textit{Open-source video generators}} \\
Hunyuan 720p      & 20.5 & 19.9 & 1.0 & 18.5 \\
Open-Sora 2.0     & 20.5 & 15.4 & 1.0 & 17.0 \\
Open-Sora-Plan 1.3 & 9.6  & 4.4  & 3.9 & 7.5  \\
CogVideoX-1.5     & 19.2 & 15.4 & 0.0 & 16.3 \\
Step-video-T2V    & 18.9 & 19.1 & 2.0 & 17.3 \\
Wanx-2.1          & 22.0 & 18.7 & 1.1 & 18.9 \\
LTX-Video         & 8.0  & 4.2  & 0.0 & 6.2  \\
Wan2.2-TI2V-5B    & 17.9 & 14.0 & 1.0 & 15.0 \\
LTX-Video-2.5     & 19.4 & 15.9 & 1.0 & 16.5 \\
\HThree           & 70.2 & 65.9 & 47.6 & 66.6 \\

\midrule
\multicolumn{5}{l}{\textit{H3-based Agentic frameworks}} \\
NEWTON            & 73.5 & 70.1 & 47.6 & 69.9 \\
Co-Director       & 69.4 & 69.8 & 33.7  & 65.8 \\
\rowcolor{OursTint}
\OurMethod        & \textbf{79.2} & \textbf{76.1} & \textbf{54.3} & \textbf{75.8} \\

\bottomrule
\end{tabular}
\end{adjustbox}
\par\endgroup

\end{table}

\begin{table}[!t]
\centering
\caption{VideoPhy-2: automatic evaluation under the NEWTON reporting protocol. }
\label{tab:05_videophy2_auto}
\begingroup
\small
\setlength{\tabcolsep}{6pt}
\renewcommand{\arraystretch}{1.15}
\begin{adjustbox}{max width=\linewidth}
\begin{tabular}{l*{6}{c}}
\toprule
Method & \multicolumn{3}{c}{All} & \multicolumn{3}{c}{Hard} \\
\cmidrule(lr){2-4}\cmidrule(lr){5-7}
 & PC $\uparrow$ & SA $\uparrow$ & Joint $\uparrow$ & PC $\uparrow$ & SA $\uparrow$ & Joint $\uparrow$ \\
\midrule
\multicolumn{7}{l}{\textit{Open-source video generators}} \\
Cosmos-Predict2.5 & 58.6 & 28.3 & 22.7 & 46.1 & 12.2 & 7.8 \\
HunyuanVideo & 73.6 & 25.3 & 22.5 & 63.3 & 8.9 & 6.7 \\
CogVideoX-5B & 69.2 & 30.2 & 25.9 & 54.4 & 9.4 & 5.0 \\
LTX-Video-2.5 & 54.7& 28.8 & 22.2& 40.7& 8.9 & 3.9 \\
LTX-Video-2B & 74.3 & 22.9 & 21.4 & 63.3 & 6.7 & 4.4 \\
Wan2.2-TI2V-5B & 56.7 & 29.6 & 24.9 & 37.8 & 10.6 & 7.2 \\
\HThree & 69.5 & 29.8 & 23.9 & 62.2 & 7.2 & 4.4 \\
\midrule
\multicolumn{7}{l}{\textit{H3-based Agentic frameworks}} \\

NEWTON & 72.5 & 29.0 & 23.2 & 71.5 & 7.8 & 6.7 \\
Co-Director & 55.8 & 30.1 & 24.9 & 38.3 & 10.0 & 5.0 \\
\rowcolor{OursTint}\OurMethod & \textbf{83.9} & \textbf{34.5} & \textbf{30.1} & \textbf{84.4} & \textbf{14.4} & \textbf{12.8} \\
\bottomrule
\end{tabular}
\end{adjustbox}
\par\endgroup

\end{table}

\subsection{Implementation Details}

PhyEvolve is training-free. All agents use Qwen3-VL-32B-Instruct, served through vLLM in BF16 with tensor parallelism of 2 and a 32K-token context window. Decoding is greedy by default. Failed validation triggers up to three replanning attempts with temperature 0.7 and top-p 0.95; if validation still fails, we revert to the last trying.

We deploy H3 through FastVideo on eight NVIDIA H100 80GB GPUs, using sequence parallelism of 8 and 50 sampling steps. Image-conditioned tasks produce 192 frames at 24 fps while preserving the input aspect ratio. Text-only tasks produce 124 frames at 24 fps at \(1344 \times 768\) resolution. For text-only tasks, the agents plan directly from the original prompt without first generating a video or an auxiliary image; H3 then generates the final video using the enhanced prompt.

\subsection{Main Results}
\label{sec:main_results}

Across all four benchmark-level metrics in Table ~\ref{tab:unified_main}, PhyEvolve outperforms all listed methods, including the agentic baselines NEWTON and Co-Director. Compared with H3, it improves by 16.1 points on PhyGenBench Average, 9.2 on PhyWorldBench Overall Both, 6.2 on VideoPhy-2 All Joint, and 28 points on StEvo-Bench Evolution Success.

\paragraph{Performance on Out-of-Sight Evolution.}As reported in Table~\ref{tab:06_stevo}, results on StEvo-Bench demonstrate PhyEvolve’s advantage in modeling physical processes that continue out of sight. PhyEvolve achieves an Evolution Success rate of 45.7\%, outperforming the H3-based agentic baselines NEWTON (29.0\%) and Co-Director (20.0\%). This advantage extends to Progress, where PhyEvolve reaches 51.2\%, surpassing both NEWTON (36.8\%) and Co-Director (42.0\%). PhyEvolve achieves the highest scores on both Observation Control (89.5\%) and Action Control (85.2\%) among all compared variants.

\paragraph{Performance on In-Sight Evolution.}The benefits extend beyond interrupted observation. On PhyGenBench (Table~\ref{tab:02_phygenbench}), the largest gains over H3 occur in Thermal and Material, at 24.4 and 23.3 points. PhyEvolve leads in Mechanics, Thermal, and Material, while NEWTON remains stronger in Optics (89.3\% versus 84.7\%). On PhyWorldBench (Table~\ref{tab:03_phyworldbench}), PhyEvolve achieves an Overall Performance score of 75.8\%, surpassing NEWTON’s 69.9\%, and outperforms NEWTON in Fundamental Physics, Composite Physics, and Anti-Physics. Its Anti-Physics Both score reaches 54.3\%, compared with 47.6\% for both H3 and NEWTON. In this setting, success means following deliberately unphysical instructions rather than conforming to real-world laws.

Results on PhyWorldBench show that PhyEvolve follows instructions effectively, even when they deliberately violate physical laws. Despite these gains on PhyWorldBench, satisfying physical plausibility and instruction fidelity jointly remains challenging on VideoPhy-2 (Table~\ref{tab:05_videophy2_auto}). PhyEvolve leads on All Joint (30.1\%) and Hard Joint (12.8\%). On Hard, its Joint score exceeds NEWTON’s 6.7\% and H3’s 4.4\%, with improvements over H3 in both physical commonsense (84.4\% vs. 62.2\%) and semantic adherence (14.4\% vs. 7.2\%). However, semantic adherence remains low, imposing a tighter ceiling on Joint success than physical commonsense under this evaluator. This gap shows that stronger physics scores alone do not ensure that the requested event is faithfully realized.

\begin{figure}[t]
    \centering
    \includegraphics[width=\linewidth]{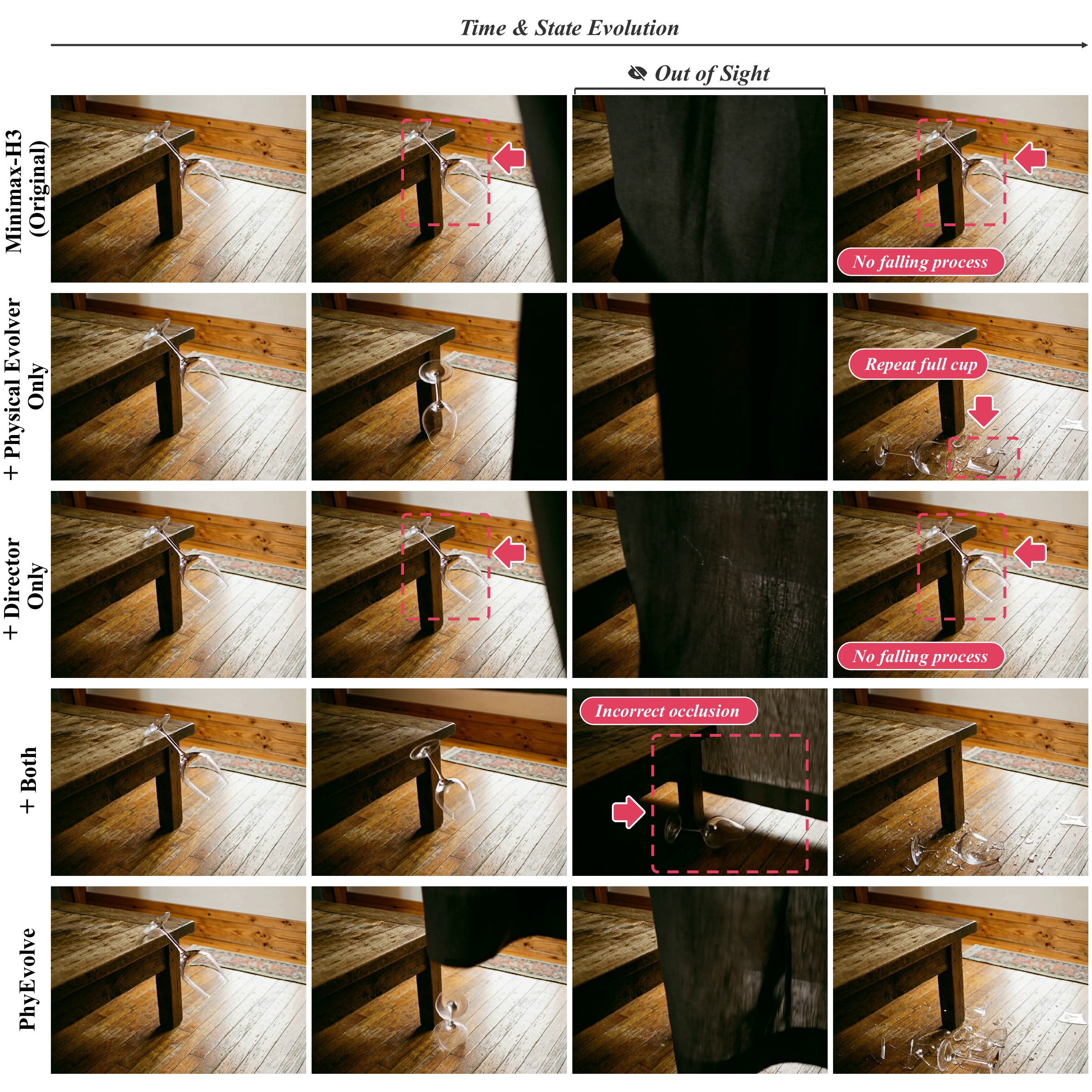}
    \caption{Component ablation on a scene where a falling glass is occluded by a curtain.}
    \label{fig:ablation_falling_glass}
\end{figure}

\begin{figure*}[!t]
    \centering
    \includegraphics[width=0.87\textwidth]{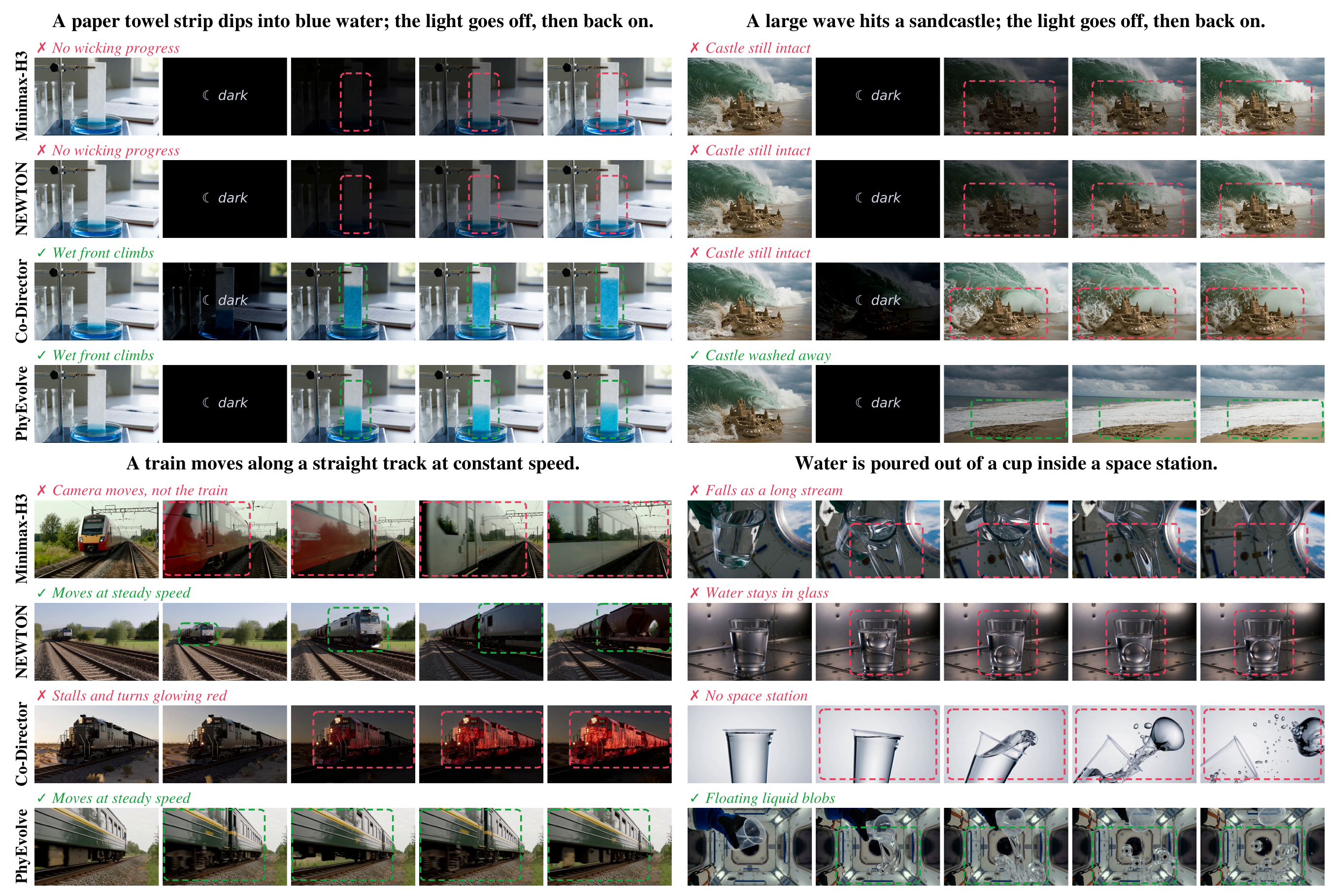}
    \caption{\textbf{Qualitative comparison on state-evolution prompts.}
    We compare Minimax-H3, NEWTON, Co-Director, and \textbf{PhyEvolve (ours)} on four cases
    from StEvo-Bench (Top)
    and other in-sight physical benchmark (Bottom).
    Frames are uniformly sampled along time from left to right; \textit{dark} marks the light-off interval.
    green boxes mark physically consistent state changes, and
    red boxes locate failures.
    }
    \label{fig:qualitative_comparison}
\end{figure*}

\begin{table}[!t]
\centering
\caption{Component ablation on StEvo-Bench.}
\label{tab:ew_component_ablation}
\begingroup
\small
\setlength{\tabcolsep}{3pt}
\renewcommand{\arraystretch}{1.1}
\begin{adjustbox}{max width=\linewidth}
\begin{tabular}{l*{6}{c}}
\toprule
Variant & \multicolumn{6}{c}{StEvo-Bench ($\%$)} \\
\cmidrule(lr){2-7}
& \shortstack{Evo.\\Success $\uparrow$}
& Prog. $\uparrow$
& Phys. $\uparrow$
& Coh. $\uparrow$
& \shortstack{Obs.\\Ctrl.$\uparrow$}
& \shortstack{Act.\\Ctrl.$\uparrow$}\\
\midrule
Baseline
  & 17.7 & 25.3 & 70.6 & 88.2 & 87.6 & 81.5 \\ 
w/o Director
  & 45.3 & 51.2 & 93.7 & 94.6 & 82.4 & 82.4 \\
w/o Physical Evolver
  & 29.0 & 38.0 & 72.3 & 94.5 & 88.4 & 78.6 \\
w/o Composer
  & 44.3 & \textbf{51.5} & 93.7 & 94.6 & 89.3 & 84.9 \\
\rowcolor{OursTint}\OurMethod
  & \textbf{45.7} & 51.2 & \textbf{96.8} & 94.6 & \textbf{89.5} & \textbf{85.2} \\
\bottomrule
\end{tabular}
\end{adjustbox}
\par\endgroup
\end{table}

\begin{table}[!t]
\centering
\caption{Agent-backbone comparison on PhyGenBench.
}
\label{tab:ew_backbone_ablation}
\begingroup
\small
\setlength{\tabcolsep}{3pt}
\renewcommand{\arraystretch}{1.1}
\begin{adjustbox}{max width=\linewidth}
\begin{tabular}{l*{5}{c}}
\toprule
Agent backbone & \multicolumn{5}{c}{PhyGenBench ($\%$)} \\
\cmidrule(lr){2-6}
& Mechanics $\uparrow$
& Optics $\uparrow$
& Thermal $\uparrow$
& Material $\uparrow$
& Average $\uparrow$ \\
\midrule
InternVL3.5-30B
  & 71.7 & \textbf{87.3} & 76.7 & 70.8 & 77.3 \\
Qwen3.5-9B
  & 67.5 & 84.0 & 81.1 & 75.8 & 77.3 \\
\rowcolor{OursTint}Qwen3-VL-32B
  & \textbf{73.3} & 84.7 & \textbf{82.2} & \textbf{77.5} & \textbf{79.6} \\
\bottomrule
\end{tabular}
\end{adjustbox}
\par\endgroup
\end{table}

\subsection{Ablation Study}

\paragraph{Component ablations.}
We compare the original-prompt baseline, full PhyEvolve, and three agent ablations on StEvo-Bench, holding H3, initial frames, and generation settings fixed.
The w/o Director variant bypasses the Director's scene grounding and action--observation planning, passing the original inputs directly to the subsequent agents.
The w/o Physical Evolver variant omits physical mechanisms analysis and causal state rollout; the Composer constructs the prompt from the original request and the Director's outputs.
The w/o Composer variant directly concatenates the Director and Physical Evolver outputs as the video-generation prompt.

Table~\ref{tab:ew_component_ablation} shows that removing Physical Evolver content produces the largest drops in Evolution, Progress, and Physics: 16.7, 13.2, and 24.5 percentage points, respectively.
Removing the Director reduces Observation Control and Action Control by 7.1 and 2.8 points, while Evolution falls by 0.4 points.
Compared with direct concatenation, the Composer improves Evolution and Physics by 1.4 and 3.1 points, but lowers Progress by 0.3 points.
In Figure~\ref{fig:ablation_falling_glass}, the H3 baseline and w/o Physical Evolver variant fail to depict the glass falling.
The w/o Director variant depicts the fall but produces an intact glass among the fragments, while w/o Composer leaves the glass visible during the intended curtain occlusion.
PhyEvolve depicts the fall, temporary occlusion, and shattered end state in sequence.

\paragraph{Agent backbones.}
We evaluate the effect of different agent backbones on PhyGenBench, as shown in Table~\ref{tab:ew_backbone_ablation}. Qwen3-VL-32B achieves the highest average score of 79.6\%, followed by InternVL3.5-30B at 77.3\% and Qwen3.5-9B at 77.3\%.

\paragraph{User Study.}
Table~\ref{tab:ew_human_evaluation} reports user preferences across four benchmarks.
PhyEvolve receives preference rates above 60\% under both Video and Physics in every comparison.
For these two criteria, respectively, the overall rates are 74.6\% and 75.3\% against direct Minimax-H3 generation, and 75.4\% and 74.7\% against NEWTON.

\paragraph{Efficiency.}
We compare LLM usage and end-to-end runtime with Minimax-H3 held fixed as the video generator (Table~\ref{tab:ew_agent_efficiency}). PhyEvolve uses 6.1 LLM calls and 17.0k tokens, the lowest values among the compared methods. Its end-to-end runtime is 452.5 seconds, compared with 1,826.5 seconds for NEWTON and 1,009.6 seconds for Co-Director, corresponding to runtime reductions of 75.2\% and 55.2\%, respectively.

\begin{table}[t]
\centering
\caption{User preference for PhyEvolve over each baseline (\%).
Each entry reports the percentage of votes favoring PhyEvolve.}
\label{tab:ew_human_evaluation}
\begingroup
\small
\setlength{\tabcolsep}{6pt}
\renewcommand{\arraystretch}{1.15}
\begin{adjustbox}{max width=\linewidth}
\begin{tabular}{lcccc}
\toprule
Benchmark
& \multicolumn{2}{c}{vs. \HThree{}}
& \multicolumn{2}{c}{vs. NEWTON} \\
\cmidrule(lr){2-3}\cmidrule(lr){4-5}
& Video & Physics & Video & Physics \\
\midrule
StEvo-Bench
  & 81.0 & 81.0 & 83.8 & 83.8 \\
PhyGenBench
  & 90.6 & 89.4 & 81.3 & 82.5 \\
PhyWorldBench
  & 60.6 & 66.9 & 69.4 & 70.6 \\
VideoPhy-2
  & 66.3 & 63.8 & 66.9 & 61.9 \\
\midrule
Overall
  & 74.6 & 75.3 & 75.4 & 74.7 \\
\bottomrule
\end{tabular}
\end{adjustbox}
\par\endgroup
\end{table}

\begin{table}[!t]
\centering
\caption{Efficiency comparison with \HThree{}.}
\label{tab:ew_agent_efficiency}
\begingroup
\small
\setlength{\tabcolsep}{6pt}
\renewcommand{\arraystretch}{1.15}
\begin{adjustbox}{max width=\linewidth}
\begin{tabular}{l*{3}{c}}
\toprule
Method & LLM calls $\downarrow$ & Tokens (k) $\downarrow$
&  End-to-end (s) $\downarrow$ \\
\midrule
NEWTON
  & 7.8 & 54.3 & 1826.5\\
Co-Director
  & 8.0 & 20.2 & 1009.6 \\

\rowcolor{OursTint}\OurMethod
  & \textbf{6.1} & \textbf{17.0} & \textbf{452.5} \\
\bottomrule
\end{tabular}
\end{adjustbox}
\par\endgroup
\end{table}

\subsection{Qualitative results.}

Figure~\ref{fig:qualitative_comparison} illustrates physical evolution under interrupted observation and visible physical processes.
In the light-off cases, H3 and NEWTON show no wicking progress, whereas Co-Director and PhyEvolve show the wet front advancing.
Only PhyEvolve depicts the sandcastle washed away after the wave and combines the space-station setting with floating liquid blobs; both NEWTON and PhyEvolve depict correct moving train.
Co-Director produces 8.0\,s StEvo-Bench videos, with frames sampled at matching relative temporal positions.

\section{Conclusion}

We presented PhyEvolve, a training-free multi-agent framework that guides video generation through structured state-evolution planning. The framework uses an explicit state trajectory to coordinate physical actions and observation timing, tracking ongoing changes even when objects are out of sight. With H3 as the fixed generator, PhyEvolve achieves the highest benchmark-level scores among the compared methods on PhyGenBench, PhyWorldBench, VideoPhy-2, and StEvo-Bench, without retraining or model modification. These results support physical-state evolution planning as a useful way to condition existing video generators and motivate further work on preserving instruction fidelity alongside state evolution.

\section*{Limitations}

PhyEvolve reasons about physical states through natural-language prompt enhancement and state-evolution planning. Its reliability depends on the agents' reasoning capabilities and available scene evidence, and it does not currently provide quantitatively calibrated predictions of physical dynamics or precise transition times. On the Hard subset of VideoPhy-2, the joint success rate for semantic adherence and physical consistency remains low, indicating that satisfying both requirements remains challenging on difficult tasks. Furthermore, our evaluation is based on short benchmark clips, and PhyEvolve's performance on longer video sequences has yet to be established. Investigating its effectiveness over longer sequences will be a key focus of our future work.


\bibliography{custom}

@article{huang2024genmac,
      title={GenMAC: Compositional Text-to-Video Generation with Multi-Agent Collaboration}, 
      author={Kaiyi Huang and Yukun Huang and Xuefei Ning and Zinan Lin and Yu Wang and Xihui Liu},
      year={2024},
      eprint={2412.04440},
      archivePrefix={arXiv},
      primaryClass={cs.CV},
      url={https://arxiv.org/abs/2412.04440}, 
}

@article{agarwal2025cosmos,
  title={Cosmos world foundation model platform for physical ai},
  author={Agarwal, Niket and Ali, Arslan and Bala, Maciej and Balaji, Yogesh and Barker, Erik and Cai, Tiffany and Chattopadhyay, Prithvijit and Chen, Yongxin and Cui, Yin and Ding, Yifan and others},
  journal={arXiv preprint arXiv:2501.03575},
  year={2025}
}

@article{yuan2026helios,
  title={Helios: Real real-time long video generation model},
  author={Yuan, Shenghai and Yin, Yuanyang and Li, Zongjian and Huang, Xinwei and Yang, Xiao and Yuan, Li},
  journal={arXiv preprint arXiv:2603.04379},
  year={2026}
}

@article{wu2026infinite,
  title={Infinite-world: Scaling interactive world models to 1000-frame horizons via pose-free hierarchical memory},
  author={Wu, Ruiqi and He, Xuanhua and Cheng, Meng and Yang, Tianyu and Zhang, Yong and Kang, Zhuoliang and Cai, Xunliang and Wei, Xiaoming and Guo, Chunle and Li, Chongyi and others},
  journal={arXiv preprint arXiv:2602.02393},
  year={2026}
}

@article{sun2025worldplay,
  title={Worldplay: Towards long-term geometric consistency for real-time interactive world modeling},
  author={Sun, Wenqiang and Zhang, Haiyu and Wang, Haoyuan and Wu, Junta and Wang, Zehan and Wang, Zhenwei and Wang, Yunhong and Zhang, Jun and Wang, Tengfei and Guo, Chunchao},
  journal={arXiv preprint arXiv:2512.14614},
  year={2025}
}

@article{huang2026solarwm,
  title={SolarWM: Open Data and Scalable Training for Long-Horizon Video World Models},
  author={Huang, Junchao and Fang, Guian and Qian, Shengju and Kong, Xianghao and Zhao, Zhuoran and Huang, Wei and Du, Yihua and Zhang, Zixin and Cui, Justin and Gu, Yuchao and others},
  journal={arXiv preprint arXiv:2609.02886},
  year={2026}
}

@article{piloto2022intuitive,
  title={Intuitive physics learning in a deep-learning model inspired by developmental psychology},
  author={Piloto, Luis S and Weinstein, Ari and Battaglia, Peter and Botvinick, Matthew},
  journal={Nature human behaviour},
  volume={6},
  number={9},
  pages={1257--1267},
  year={2022},
  publisher={Nature Publishing Group UK London}
}

@inproceedings{hafner2019learning,
  title={Learning latent dynamics for planning from pixels},
  author={Hafner, Danijar and Lillicrap, Timothy and Fischer, Ian and Villegas, Ruben and Ha, David and Lee, Honglak and Davidson, James},
  booktitle={International conference on machine learning},
  pages={2555--2565},
  year={2019},
  organization={PMLR}
}

@inproceedings{stapf2026composition,
  title={Composition of memory experts for diffusion world models},
  author={Stapf, Sebastian and Acuaviva, Pablo and Davtyan, Aram and Favaro, Paolo},
  booktitle={International Conference on Learning Representations},
  volume={2026},
  pages={15377--15403},
  year={2026}
}

@article{xu2026teaching,
  title={Teaching Video Generators to Remember: Eliciting Dynamic Memory for Out-of-Sight State Evolution},
  author={Xu, Tianshuo and Xie, Yichen and Meng, Depu and Peng, Chensheng and Herau, Quentin and Jiang, Bo and Hu, Yihan and Zhan, Wei},
  journal={arXiv preprint arXiv:2605.25333},
  year={2026}
}

@article{chen2026out,
  title={Out of sight but not out of mind: Hybrid memory for dynamic video world models},
  author={Chen, Kaijin and Liang, Dingkang and Zhou, Xin and Ding, Yikang and Liu, Xiaoqiang and Wan, Pengfei and Bai, Xiang},
  journal={arXiv preprint arXiv:2603.25716},
  year={2026}
}

@inproceedings{huang2024vbench,
  title={Vbench: Comprehensive benchmark suite for video generative models},
  author={Huang, Ziqi and He, Yinan and Yu, Jiashuo and Zhang, Fan and Si, Chenyang and Jiang, Yuming and Zhang, Yuanhan and Wu, Tianxing and Jin, Qingyang and Chanpaisit, Nattapol and others},
  booktitle={2024 IEEE/CVF Conference on Computer Vision and Pattern Recognition (CVPR)},
  pages={21807--21818},
  year={2024},
  organization={IEEE}
}

@inproceedings{sun2025t2v,
  title={T2v-compbench: A comprehensive benchmark for compositional text-to-video generation},
  author={Sun, Kaiyue and Huang, Kaiyi and Liu, Xian and Wu, Yue and Xu, Zihan and Li, Zhenguo and Liu, Xihui},
  booktitle={2025 IEEE/CVF Conference on Computer Vision and Pattern Recognition (CVPR)},
  pages={8406--8416},
  year={2025},
  organization={IEEE}
}

@article{meng2024towards,
  title={Towards world simulator: Crafting physical commonsense-based benchmark for video generation},
  author={Meng, Fanqing and Liao, Jiaqi and Tan, Xinyu and Shao, Wenqi and Lu, Quanfeng and Zhang, Kaipeng and Cheng, Yu and Li, Dianqi and Qiao, Yu and Luo, Ping},
  journal={arXiv preprint arXiv:2410.05363},
  year={2024}
}

@inproceedings{gu2026phyworldbench,
  title={$ PhyWorldBench $: A Comprehensive Evaluation of Physical Realism in Text-to-Video Models},
  author={Gu, Jing and Liu, Xian and Zeng, Yu and Nagarajan, Ashwin and Zhu, Fangrui and Hong, Daniel and Fan, Yue and Yan, Qianqi and Zhou, Kaiwen and Liu, Ming-Yu and others},
  booktitle={International Conference on Learning Representations},
  volume={2026},
  pages={75130--75164},
  year={2026}
}

@inproceedings{bansal2025videophy,
  title={Videophy: Evaluating physical commonsense for video generation},
  author={Bansal, Hritik and Lin, Zongyu and Xie, Tianyi and Zong, Zeshun and Yarom, Michal and Bitton, Yonatan and Jiang, Chenfanfu and Sun, Yizhou and Chang, Kai-Wei and Grover, Aditya},
  booktitle={International Conference on Learning Representations},
  volume={2025},
  pages={102075--102121},
  year={2025}
}

@inproceedings{bansal2026videophy,
  title={Videophy-2: A challenging action-centric physical commonsense evaluation in video generation},
  author={Bansal, Hritik and Peng, Clark and Bitton, Yonatan and Goldenberg, Roman and Grover, Aditya and Chang, Kai-Wei},
  booktitle={International Conference on Learning Representations},
  volume={2026},
  pages={118456--118470},
  year={2026}
}

@inproceedings{motamed2026generative,
  title={Do generative video models understand physical principles?},
  author={Motamed, Saman and Culp, Laura and Swersky, Kevin and Jaini, Priyank and Geirhos, Robert},
  booktitle={2026 IEEE/CVF Winter Conference on Applications of Computer Vision (WACV)},
  pages={948--958},
  year={2026},
  organization={IEEE}
}

@article{tragoudaras2025evaluating,
  title={Evaluating Newtonian Mechanics in Video Generative Models with Real Physical Systems},
  author={Tragoudaras, Antonios and Zhang, Chenyu and Cherniavskii, Daniil and Vozikis, Antonios and Nijdam, Thijmen and Prinzhorn, Derck WE and Bodracska, Mark and Sebe, Nicu and Zadaianchuk, Andrii and Gavves, Efstratios},
  journal={arXiv preprint arXiv:2504.02918},
  year={2025}
}

@article{li2026learning,
  title={Learning Explicit Physical Parameter Control and Benchmarking for Video Generation},
  author={Li, Yanxun and Wen, Hao and Song, Bingze and Zhu, Jiashu and Hao, Aiming and Chen, Chubin and Chen, Jintao and Wu, Jiahong and Chu, Xiangxiang and Wang, Miao},
  journal={arXiv preprint arXiv:2607.18924},
  year={2026}
}

@inproceedings{ma2026out,
  title={Out of sight, out of mind? evaluating state evolution in video world models},
  author={Ma, Ziqi and Liufu, Mengzhan and Gkioxari, Georgia},
  booktitle={European Conference on Computer Vision},
  pages={129--146},
  year={2026},
  organization={Springer}
}

@article{lin2023videodirectorgpt,
  title={Videodirectorgpt: Consistent multi-scene video generation via llm-guided planning},
  author={Lin, Han and Zala, Abhay and Cho, Jaemin and Bansal, Mohit},
  journal={arXiv preprint arXiv:2309.15091},
  year={2023}
}

@article{chen2026code,
  title={Code world model: Coding agent as world brain},
  author={Chen, Yiwen and Lin, Guosheng and Zhang, Chi},
  journal={arXiv preprint arXiv:2608.25927},
  year={2026}
}

@article{cudlenco2026agentic,
  title={Agentic Video Generation: From Text to Executable Event Graphs via Tool-Constrained LLM Planning},
  author={Cudlenco, Nicolae and Masala, Mihai and Leordeanu, Marius},
  journal={arXiv e-prints},
  pages={arXiv--2604},
  year={2026}
}

@article{yuan2024mora,
  title={Mora: Enabling generalist video generation via a multi-agent framework},
  author={Yuan, Zhengqing and Liu, Yixin and Cao, Yihan and Sun, Weixiang and Jia, Haolong and Chen, Ruoxi and Li, Zhaoxu and Lin, Bin and Yuan, Li and He, Lifang and others},
  journal={arXiv preprint arXiv:2403.13248},
  year={2024}
}

@article{hu2024storyagent,
  title={Storyagent: Customized storytelling video generation via multi-agent collaboration},
  author={Hu, Panwen and Jiang, Jin and Chen, Jianqi and Han, Mingfei and Liao, Shengcai and Chang, Xiaojun and Liang, Xiaodan},
  journal={arXiv preprint arXiv:2411.04925},
  year={2024}
}

@inproceedings{wang2026mavis,
  title={MAViS: A multi-agent framework for long-sequence video storytelling},
  author={Wang, Qian and Huang, Ziqi and Jia, Ruoxi and Debevec, Paul and Yu, Ning},
  booktitle={Proceedings of the 19th Conference of the European Chapter of the Association for Computational Linguistics (Volume 1: Long Papers)},
  pages={2273--2295},
  year={2026}
}

@misc{song2026codirector,
      title={Co-Director: Agentic Generative Video Storytelling}, 
      author={Yale Song and Yiwen Song and Nick Losier and Nathan Hodson and Ye Jin and Rhyard Zhu and Yan Xu and Daniel Vlasic and Carina Claassen and Jasmine Leon and Khanh G. LeViet and Zack Chomyn and Joe Timmons and Brett Slatkin and Scott Penberthy and Tomas Pfister},
      year={2026},
      eprint={2604.24842},
      archivePrefix={arXiv},
      primaryClass={cs.AI},
      url={https://arxiv.org/abs/2604.24842}, 
}

@inproceedings{xue2025phyt2v,
  title={Phyt2v: Llm-guided iterative self-refinement for physics-grounded text-to-video generation},
  author={Xue, Qiyao and Yin, Xiangyu and Yang, Boyuan and Gao, Wei},
  booktitle={2025 IEEE/CVF Conference on Computer Vision and Pattern Recognition (CVPR)},
  pages={18826--18836},
  year={2025},
  organization={IEEE}
}

@article{feng2026newton,
  title={NEWTON: Agentic Planning for Physically Grounded Video Generation},
  author={Feng, Yuxiang and Wang, Juncheng and Xu, Chao and Qian, Yijie and Wang, Huihan and Hou, Wenlong and Liu, Yang and Sun, Baigui and Liu, Yong and Wang, Shujun},
  journal={arXiv preprint arXiv:2605.18396},
  year={2026}
}

@inproceedings{bai2026moregen,
  title={Moregen: Multi-agent motion-reasoning engine for code-based text-to-video synthesis},
  author={Bai, Xiangyu and Liang, He and Galoaa, Bishoy and Nandi, Utsav and Moezzi, Shayda and He, Yuhang and Ostadabbas, Sarah},
  booktitle={Proceedings of the IEEE/CVF Conference on Computer Vision and Pattern Recognition},
  pages={7632--7642},
  year={2026}
}

@article{wang2026physically,
  title={Physically Plausible Video Generation via Visual-Semantic Chain-of-Events Conditioning},
  author={Wang, Zixuan and Hu, Yixin and Li, Wen and Chen, Feng and Liu, Yan and Peng, Duo and Lei, Yinjie},
  journal={arXiv preprint arXiv:2609.00656},
  year={2026}
}

@article{minimaxh32026,
  title={Minimax h3: An open model breaking the boundaries between tasks and modalities},
  author={MiniMax},
  url={https://www.minimax.io/blog/minimax-h3},
  year={2026}
}

\appendix
\newpage

\section{Appendix}
\label{sec:appendix}

\subsection{Agent Prompt}
Figures~\ref{fig:prompt_director}, \ref{fig:prompt_physical_evolver}, and~\ref{fig:prompt_composer} present the role-level prompt templates for the Director, Physical Evolver, and Composer, respectively. Each template groups the task instructions, output requirements, and review procedure for its role; a role may involve multiple separate model calls.

\paragraph{Director.}
The prompt in Figure~\ref{fig:prompt_director} instructs the Director to ground the scene in the user request and optional initial frame, and to distinguish physical events from observation changes. Its outputs record the initial entities and states, an ordered shot plan, and the permitted causal effects of each event. Review instructions check source coverage, scene grounding, and event classification, with revisions restricted to flagged fields.

\paragraph{Physical Evolver.}
The prompt in Figure~\ref{fig:prompt_physical_evolver} separates physical mechanisms analysis from causal state rollout. Given the original request, available scene evidence, and the Director's records, the agent first identifies physical mechanisms, their activation conditions, and immediate responses. It then propagates the reviewed records into dependent state transitions, including changes that continue during unobserved intervals. Mechanisms and transitions are reviewed separately to check prerequisites, responses, and state dependencies while retaining uncertainty where the inputs do not determine an outcome.

\paragraph{Composer.}
The prompt in Figure~\ref{fig:prompt_composer} instructs the Composer to integrate the original request with the reviewed scene and physical records into a complete replacement prompt. It preserves the requested actions, event order, and camera behavior while expressing the planned physical conditions and state changes. A separate fidelity review then compares the draft with the upstream records. Revisions are restricted to flagged sentences to correct omissions or contradictions while preserving unrelated text.

\begin{figure*}[t]
    \centering
    \setlength{\textfloatsep}{6pt}
    \begin{systemprompt}
    \small
    \textbf{Director: Scene Grounding and Action--Observation Disentanglement}\\[4pt]
    \textbf{Role and inputs.}
    You are the Director. Given the original video request and an initial frame
    when available, establish the initial scene, an ordered shot plan, and the
    permitted causal roles of source events. Preserve the requested subjects,
    actions, event order, and camera behavior.
    \par\smallskip
    \textbf{Module 1: Scene grounding.}
    Extract exact source clauses and identify the entities, initial properties,
    positions, and contacts supported by the inputs. Distinguish textual evidence
    from visible evidence; do not claim visual observations without an image.
    Treat the initial scene as given, without explaining how it arose.
    Do not infer zero velocity from a still frame or invent an invisible holder,
    support, or release. Retain explicitly requested actors and actions even when
    they are not visible initially. Construct a concise ordered shot plan without
    predicting future physical outcomes.
    \par\smallskip
    \textbf{Module 2: Action--Observation Disentanglement.}
    Classify each relevant source event as an observation change, physical action,
    environment change, or scene constraint. Associate it with an exact source
    quote and state what it is allowed to change. Separate changes in visibility
    from physical drivers. Occlusion, camera movement, and illumination used only
    to conceal the scene do not start, stop, reset, or release physical processes.
    Preserve genuine source-grounded causal effects; do not discard a physical
    action merely because it coincides with concealment. Use event order and
    prerequisites rather than invented numerical timestamps.
    \par\smallskip
    \textbf{Structured outputs.}
    The scene plan contains \texttt{source\_clauses},
    \texttt{initial\_entities} (entity, state, evidence), and \texttt{shot\_plan}.
    The action--observation contract contains \texttt{events}
    (source quote, kind, causal role) and \texttt{constraints}.
    Keep both records concise and source-grounded. Do not produce the final
    video prompt or supply a guessed physical outcome in place of a source action.
    \par\smallskip
    \textbf{Separate review and local refinement.}
    The reviewer receives the original request, available initial-frame evidence,
    and the candidate plan and contract. Check source coverage, scene grounding,
    shot order, and causal classification. Report concrete defects using existing
    candidate field paths, supporting evidence, and minimal correction instructions.
    As the Director, revise only the approved fields; preserve all unrelated
    content and the output structure. The controller validates the patch and
    submits the revised candidate for review. Allow at most three local revisions;
    if review remains unresolved or fails, retain the latest structurally valid
    candidate and record the unresolved status without claiming acceptance.
    \end{systemprompt}
    \caption{\textbf{Role-level prompt template for the Director.}
    Scene grounding establishes the initial conditions, while action--observation
    disentanglement separates physical drivers from visibility changes.
    A separate reviewer guides bounded local revisions of the structured plan.}
    \label{fig:prompt_director}
\end{figure*}

\begin{figure*}[t]
    \centering
    \setlength{\textfloatsep}{6pt}
    \begin{systemprompt}
    \small
    \textbf{Physical Evolver: Physical Mechanisms Analysis and Causal State Rollout}\\[4pt]
    \textbf{Role and inputs.}
    You are the Physical Evolver. Use the original request, available initial-frame
    evidence, and the Director's scene plan and action--observation contract to
    infer physical evolution. Produce physical records, not camera instructions,
    stylistic descriptions, or a complete video prompt.
    \par\smallskip
    \textbf{Module 1: Physical Mechanisms Analysis.}
    Identify concise local mechanisms, including those implicit in the initial
    scene. For each mechanism, record the affected entities, evidence, actual
    prerequisite, mechanism, and immediate observable response. Distinguish
    supplied observations from inferred physical responses. At this stage, keep
    facts independent: do not propagate intermediate states or construct a global
    before/during/after-occlusion narrative.
    \par\smallskip
    \textbf{Module 2: Causal State Rollout.}
    Starting from the initial scene, propagate the identified mechanisms into
    ordered state transitions. Each transition specifies its preceding state,
    actual prerequisite, observable change, resulting state, supporting fact IDs,
    and dependencies on earlier transitions. Carry the relevant resulting state
    forward; independent transitions may occur concurrently. Continue applicable
    mechanisms while objects are hidden, and preserve the evolved state when
    visibility returns. Visibility changes alone must not start or reset a
    process. Do not add mechanisms absent from the supplied facts or force a
    final equilibrium within the clip.
    \par\smallskip
    \textbf{Physical constraints.}
    Treat the initial scene as given. Do not invent hidden supports, holders,
    external actors, or a required release event. Ordinary gravity is a physical
    default, not an unsupported hidden force; account for different conditions
    only when supported by the inputs. Underwater, gravity and buoyancy coexist.
    A still image does not establish zero velocity. Do not assume friction is
    sufficient to arrest motion without evidence. Preserve genuine prerequisites,
    source actions, and uncertainty; do not invent numerical timing or accelerate
    a naturally gradual process to guarantee completion.
    \par\smallskip
    \textbf{Structured outputs.}
    Return local \texttt{facts} with fields \texttt{id}, \texttt{entities},
    \texttt{evidence}, \texttt{condition}, \texttt{mechanism}, and \texttt{response}.
    Return rollout \texttt{transitions} with fields \texttt{id},
    \texttt{fact\_ids}, \texttt{entities}, \texttt{before}, \texttt{condition},
    \texttt{change}, \texttt{after}, \texttt{depends\_on}, and \texttt{evidence}.
    \par\smallskip
    \textbf{Separate review and local refinement.}
   Jointly review local mechanisms and state transitions in one separate call. The reviewer receives the original request, relevant evidence, upstream records, and the complete candidate output. Check prerequisites, responses, state dependencies, and consistency between mechanisms and transitions. Return evidence-grounded defects at existing field paths. As the Physical Evolver, modify only approved fields and preserve unrelated entries; upstream records remain read-only. Validate each patch and jointly review the combined records again, with at most three local revisions. Retain the latest structurally valid candidate if review is unresolved or fails, and record that status separately from physical correctness.
    \end{systemprompt}
    \caption{\textbf{Role-level prompt template for the Physical Evolver.}
    Physical Mechanisms Analysis establishes evidence, prerequisites, and
    responses; causal state rollout propagates them into continuous physical
    evolution. Both steps share one generation call, followed by a joint review of their outputs and local revisions when needed.}
    \label{fig:prompt_physical_evolver}
\end{figure*}

\begin{figure*}[t]
    \centering
    \setlength{\textfloatsep}{6pt}
    \begin{systemprompt}
    \small
    \textbf{Composer: Prompt Construction and Physical-Fidelity Refinement}\\[4pt]
    \textbf{Role and inputs.}
    You are the Composer. Given the original request, the Director's scene plan
    and action--observation contract, and the Physical Evolver's mechanism and
    state-transition records, construct a complete replacement English video
    prompt. Use the upstream scene representation; do not invent additional
    observations or physical mechanisms.
    \par\smallskip
    \textbf{Module 1: Complete prompt construction.}
    Integrate the requested scene, subjects, actions, camera behavior, and supplied
    physical evolution into concise continuous prose. Preserve physical
    prerequisites, transition dependencies, and the state reached after hidden
    evolution. Express observation changes without making them unauthorized
    physical triggers. Keep the original intent and event order. Replace the
    original short prompt rather than appending an augmentation to it; avoid
    repeated sentences and disconnected instructions.
    \par\smallskip
    \textbf{Fidelity and pacing constraints.}
    Express only supplied physical responses and transitions. Do not invent
    additional actors, mechanisms, or outcomes, and do not construct missing
    transition chains. When only independent facts are supplied, integrate them
    fluently while retaining explicit source order. Do not introduce numerical
    timestamps or completion deadlines. Use photorealistic footage with natural
    real-time playback unless altered playback is explicitly requested in the
    original input. Preserve natural acceleration, deceleration, and gradual
    physical changes; real-time playback does not imply uniform motion.
    \par\smallskip
    \textbf{Module 2: Physical-Fidelity Refinement --- separate review.}
    Provide the reviewer with the original request, all supplied planning and
    physical records, and the candidate split into identified sentences. The
    reviewer checks for omissions, contradictions, duplication, unauthorized
    playback changes, and invented actions or mechanisms. Each issue identifies
    the affected sentence IDs, a concrete mismatch, supporting input evidence,
    and a minimal correction instruction. For an omission, identify an existing
    sentence as the insertion point. Do not demand new physical reasoning,
    invisible supports, a release event, or a final equilibrium. Stylistic
    preferences alone are not defects.
    \par\smallskip
    \textbf{Local revision and output.}
    Return the initial prompt as a JSON object with a single
    \texttt{scene\_prompt} field; its value is continuous English prose.
    When reviewer feedback is supplied, return edits containing only approved
    \texttt{sentence\_id} and \texttt{replacement} pairs. Revise each approved
    sentence once and leave every unlisted span unchanged. The controller applies
    and validates the patch, then requests another review. Allow at most three
    local revisions. If review remains unresolved or fails, preserve the latest
    valid prompt and report its review status separately; producing a usable
    prompt does not certify its physical correctness. Deliver the final
    replacement prompt to the video generator, together with the initial frame
    when provided.
    \end{systemprompt}
    \caption{\textbf{Role-level prompt template for the Composer.}
    Prompt construction integrates the source requirements and upstream physical
    records. Physical-fidelity refinement uses evidence-grounded sentence-level
    feedback to correct omissions and contradictions while locking unrelated text.}
    \label{fig:prompt_composer}
\end{figure*}

\subsection{Additional qualitative results}

\begin{figure*}[t]
    \centering
    \includegraphics[width=\textwidth]{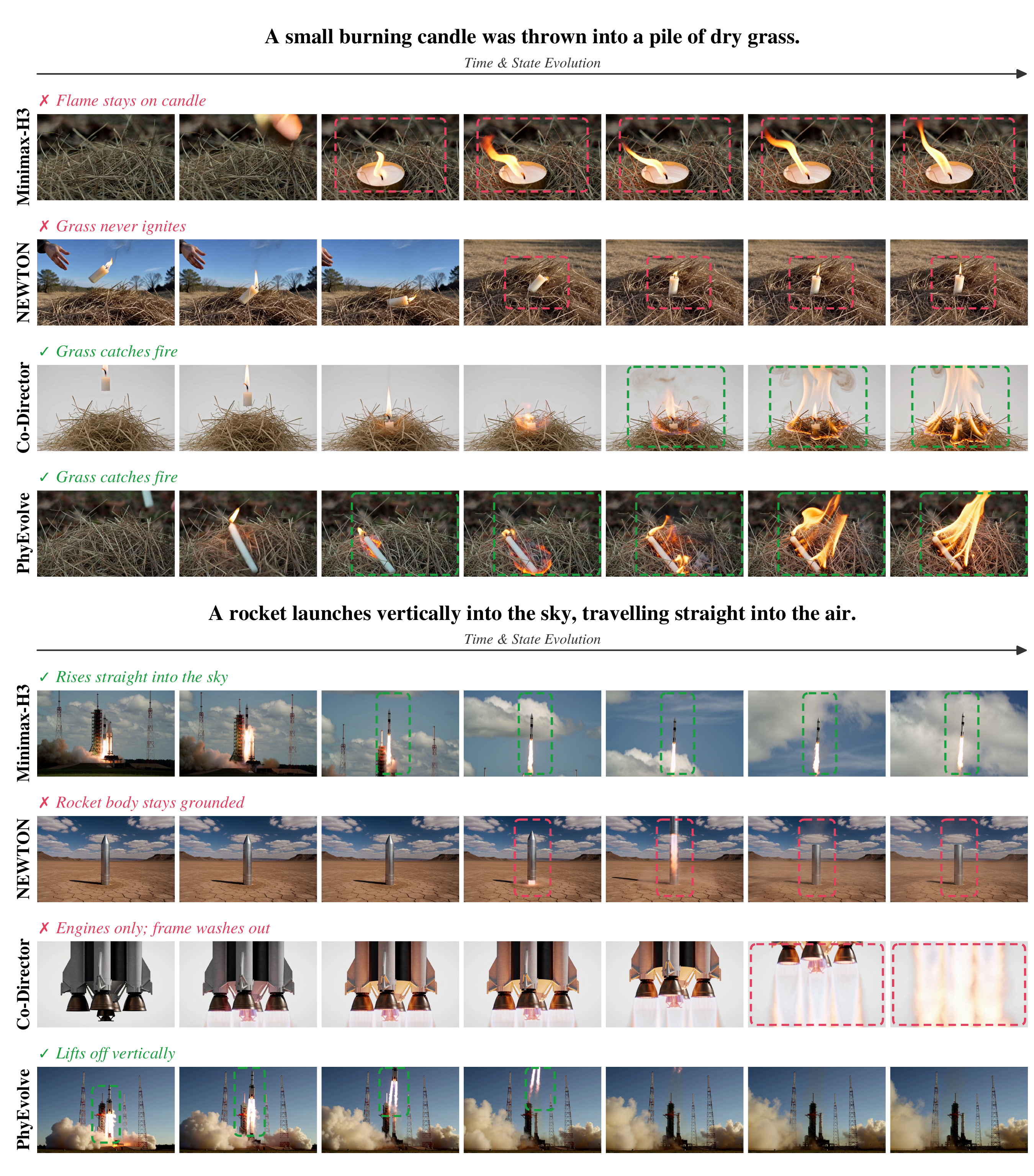}
    \caption{\textbf{Additional qualitative comparisons.}
    Top: a burning candle thrown onto dry grass . Bottom: a vertical rocket launch .
    Seven frames are sampled uniformly over time;
    green and red boxes mark correct and failed state evolution.
    Minimax-H3 and NEWTON fail to ignite the grass, and NEWTON and Co-Director fail to show a rocket leaving the ground,
    while PhyEvolve completes both processes.}
    \label{fig:appendix_fire_rocket}
\end{figure*}

We provide additional visualizations to complement the quantitative comparisons in Section~\ref{sec:main_results}. Figure~\ref{fig:appendix_fire_rocket} compares Minimax-H3, NEWTON, Co-Director, and PhyEvolve in two scenarios: a burning candle thrown onto dry grass and a vertical rocket launch. These sequences show the progression of the requested events and their resulting state changes, including grass ignition and rocket ascent.

\begin{figure*}[t]
    \centering
    \includegraphics[width=\textwidth]{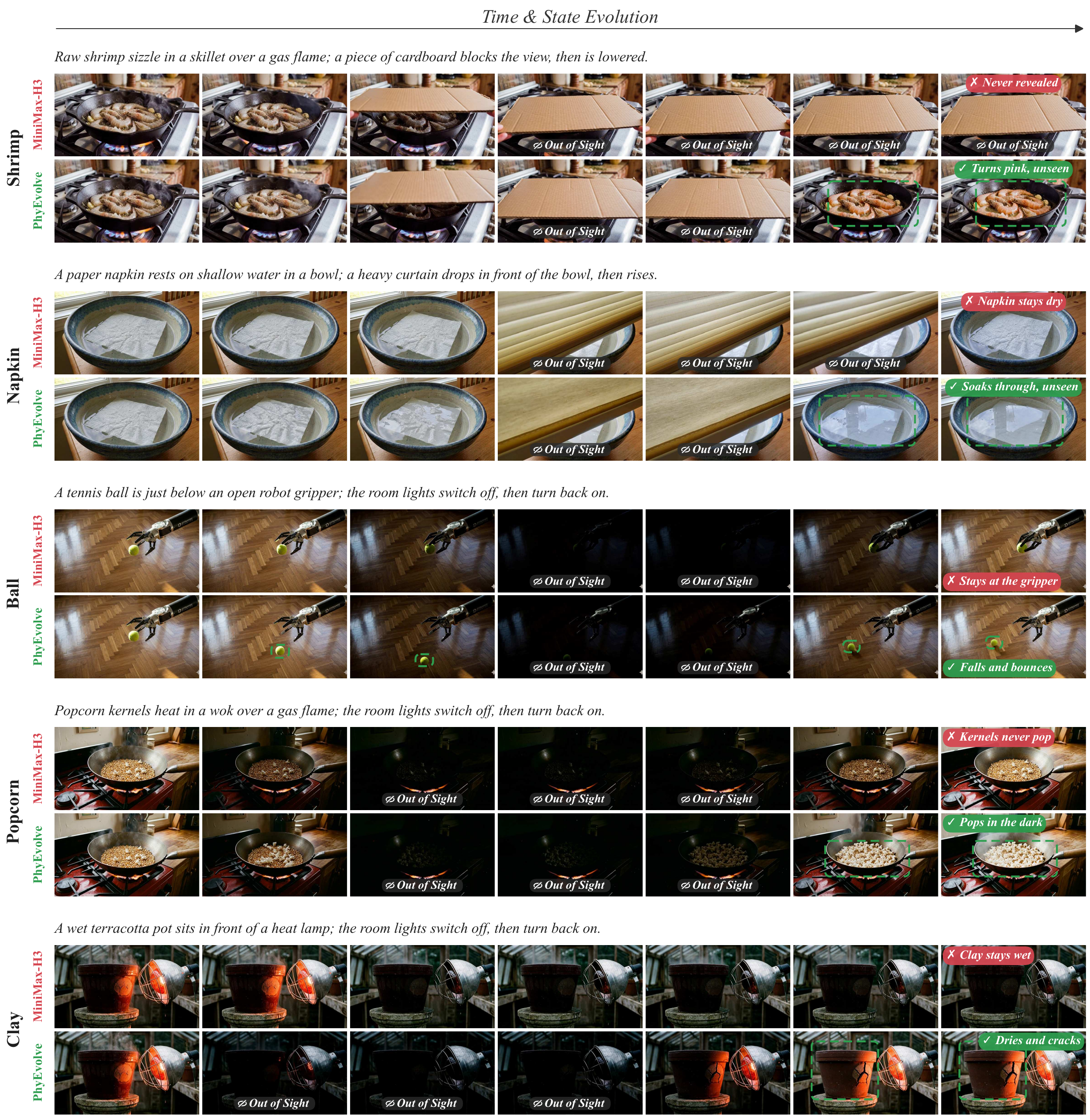}
    \caption{\textbf{Additional qualitative results on StEvo-Bench.}
    Each case contains an interval where the scene is unobserved (Lights off, curtain drop, cardboard occlusion). PhyEvolve carries the underlying state change through this interval, so the scene reappears in a physically consistent, updated state.}
    \label{fig:appendix_stevo}
\end{figure*}

\begin{figure*}[t]
    \centering
    \includegraphics[width=\textwidth]{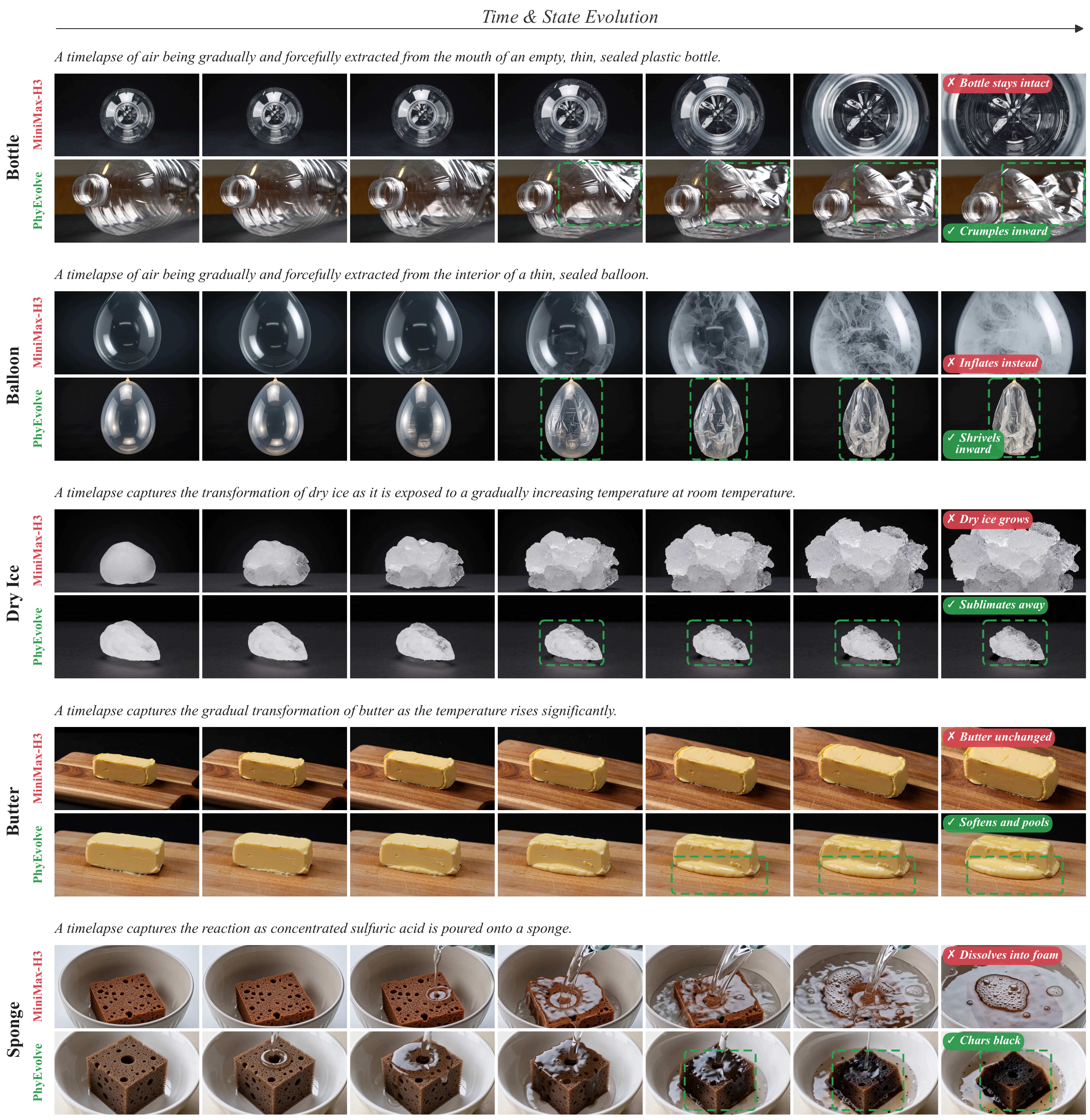}
    \caption{\textbf{Additional qualitative results on PhyGenBench.}}
    \label{fig:appendix_phygen}
\end{figure*}

Figure~\ref{fig:appendix_stevo} compares Minimax-H3 and PhyEvolve on five StEvo-Bench cases involving cardboard occlusion, curtain occlusion, or lighting changes. The cases illustrate whether the requested physical changes are reflected in object states when visibility returns, as well as failures to reveal the scene as instructed. Figure~\ref{fig:appendix_phygen} extends the comparison to PhyGenBench time-lapse prompts involving bottle deformation, balloon deflation, dry ice sublimation, butter melting, and a sponge reacting to acid. These selected examples support visual inspection of event progression and state continuity, while aggregate performance is summarized by the benchmark-level metrics in the main experiments.

\end{document}